\documentclass[11pt,a4paper]{article} % ArXiv 推荐 11pt 单栏
\usepackage{arxiv}                    % 加载 arxiv 模板样式
\renewcommand{\headeright}{}  % 清空右上角页眉
\renewcommand{\undertitle}{}  % 清空第一页标题下的文字
\usepackage{times}          % 字体
\usepackage{graphicx}       % 插图
\usepackage{amsmath}        % 数学公式
\usepackage{amssymb}        % 数学符号
\usepackage{booktabs}       % 三线表 (\toprule, \midrule)
\usepackage[numbers,sort&compress]{natbib} % 参考文献 [1-3]
\usepackage[dvipsnames]{xcolor} % 颜色支持
\usepackage{caption}        % 图表标题控制
\usepackage{subcaption}     % 子图支持
\usepackage{url}            % 网址格式
\usepackage{algorithm}      % 算法伪代码
\usepackage{algpseudocode}  % 算法伪代码

\usepackage{xspace}
\makeatletter
\DeclareRobustCommand\onedot{\futurelet\@let@token\@onedot}
\def\@onedot{\ifx\@let@token.\else.\null\fi\xspace}

\makeatother

\definecolor{cvprblue}{rgb}{0.21,0.49,0.74}
\usepackage[pagebackref,breaklinks,colorlinks,allcolors=cvprblue]{hyperref}

\usepackage{pifont}       % \ding{xx}
\usepackage{bbding}       % \Checkmark,\XSolid,... (需要和pifont宏包共同使用)
\usepackage{dsfont} % for \mathds
\usepackage[table]{xcolor} % for \rowcolor
\usepackage{enumitem}
\usepackage{rotating}
\usepackage{bbm}
\usepackage{cases}
\usepackage{caption}
\usepackage{bbding}

\usepackage{multicol}
\usepackage{subcaption}
\usepackage{wrapfig}
\usepackage{pifont}
\usepackage{amssymb}
\usepackage{listings}
\usepackage{mathtools}
\usepackage{makecell}
\usepackage{multirow}
\usepackage{graphicx}
\usepackage{booktabs}
\usepackage{colortbl}
\usepackage{amsmath} 
\usepackage[utf8]{inputenc}
\usepackage[T1]{fontenc}
\usepackage{xcolor}
\usepackage{tcolorbox}
\tcbuselibrary{breakable}
\usepackage{url}
\newcommand{\cmark}{\ding{51}}
\newcommand{\xmark}{\ding{55}}
\newcommand{\circnum}[1]{\ding{\the\numexpr 171+(#1)\relax}}

\definecolor{Gray}{gray}{0.95}

\newcommand{\gr}{\rowcolor[gray]{.8}}

\newcommand{\red}[1]{{\color{red}#1}}

\usepackage[table,xcdraw]{xcolor}
\usepackage{tcolorbox}
\tcbuselibrary{skins, breakable, listings}
\usepackage{fontawesome5} % 用于精美图标
\usepackage{enumitem}     % 用于自定义列表
\usepackage{listings}     % 代码块

\definecolor{darkblue}{RGB}{0, 51, 102}
\definecolor{lightblue}{RGB}{235, 242, 250}
\definecolor{codebg}{RGB}{248, 248, 248}
\definecolor{rulecolor}{RGB}{180, 50, 50}
\definecolor{cmdcolor}{RGB}{0, 100, 180}
\usepackage{listings}
\usepackage{xcolor}

\definecolor{codebg}{RGB}{250, 250, 250}    % 极淡的灰色背景
\definecolor{codekey}{RGB}{0, 51, 179}      % 深蓝色关键字
\definecolor{codestr}{RGB}{163, 21, 21}     % 暗红色字符串
\definecolor{codecom}{RGB}{0, 128, 0}       % 深绿色注释
\definecolor{codenum}{RGB}{128, 128, 128}   % 灰色行号
\definecolor{codeframe}{RGB}{220, 220, 220} % 浅灰色边框

\lstdefinestyle{mypython}{
    language=Python,
    backgroundcolor=\color{codebg},   
    basicstyle=\ttfamily\footnotesize\color{black}, % 字号设为 footnotesize
    keywordstyle=\color{codekey}\bfseries,  % 关键字加粗
    stringstyle=\color{codestr},            % 字符串
    commentstyle=\color{codecom}\itshape,   % 注释斜体
    tabsize=4,
    captionpos=b,
    breaklines=true,                        % 自动换行
    breakatwhitespace=false,
    showstringspaces=false,                 % 不显示字符串中的空格符号
    keepspaces=true,                        % 保持代码缩进
    frame=lines,                            % 仅显示上下边框 (看起来更像学术表格)
    framerule=1pt,                          % 边框线宽
    rulecolor=\color{codeframe},            % 边框颜色
    framexleftmargin=0.5em,                   % 背景色向左延伸，覆盖行号区
    aboveskip=1em,                          % 代码块上方的垂直间距
    belowskip=1em,                          % 代码块下方的垂直间距
}

\newtcolorbox{promptbox}[1]{
    enhanced,
    breakable,
    skin=bicolor,
    colback=white,
    colbacklower=lightblue,
    colframe=darkblue,
    title={#1},
    fonttitle=\bfseries\small, % <--- 【关键修改】这里直接设为 \small
    coltitle=white,
    attach boxed title to top left={xshift=10pt, yshift*=-10pt},
    boxed title style={
        colback=darkblue, 
        frame hidden, 
        rounded corners, 
        top=3pt, bottom=3pt, % <--- 【关键修改】调小上下内边距，让盒子更精致
        left=3pt, right=3pt
    },
    boxrule=1pt,
    left=10pt, right=10pt, top=15pt, bottom=10pt,
    shadow={2mm}{-2mm}{0mm}{black!10},
    bottomrule at break=1pt,
    toprule at break=1pt,
    pad at break=2mm,
}

\title{Automatic Attack Discovery for Few-Shot Class-Incremental Learning via Large Language Models}

\author{
  Haidong Kang\thanks{Corresponding author} \\
  School of Computer and Communication Engineering\\
  Northeastern University \\
  \texttt{hdkang@stumail.neu.edu.cn} \\
  \And
  Wei Wu \\
  School of Information and Communication\\
  University of Electronic Science and Technology of China \\
  \texttt{202522010418@std.uestc.edu.cn} \\
  \And
  Hanling Wang \\
  Department of Strategic and Advanced Interdisciplinary Research\\
  Pengcheng Laboratory \\
  \texttt{wanghl03@pcl.ac.cn} \\
}

\begin{document}

\maketitle

% [重要] ArXiv 模板需要显式的 abstract 环境
\begin{abstract}
Few-shot class incremental learning (FSCIL) is a more realistic and challenging paradigm in continual learning to incrementally learn unseen classes and overcome catastrophic forgetting on base classes with only a few training examples. Previous efforts have primarily centered around studying more effective FSCIL approaches. By contrast, less attention was devoted to thinking the security issues in contributing to FSCIL. This paper aims to provide a holistic study of the impact of attacks on FSCIL. We first derive insights by systematically exploring how human expert-designed attack methods (i.e., PGD, FGSM) affect FSCIL. 
We find that those methods either fail to attack base classes, 
or suffer from huge labor costs due to relying on huge expert knowledge. This highlights the need to craft a specialized attack method for FSCIL. Grounded in these insights, in this paper, we propose a simple yet effective ACraft method to automatically steer and discover optimal attack methods targeted at FSCIL by leveraging Large Language Models (LLMs) without human experts. Moreover, to improve the reasoning between LLMs and FSCIL, we introduce a novel Proximal Policy Optimization (PPO) based reinforcement learning to optimize learning, making LLMs generate better attack methods in the next generation by establishing positive feedback. Experiments on mainstream benchmarks show that our ACraft significantly degrades the performance of state-of-the-art FSCIL methods and dramatically beyond human expert-designed attack methods while maintaining the lowest costs of attack.

% 提出的LLM驱动攻击生成框架可推广至其他安全敏感场景（如联邦学习、少样本检测等）本工作有望成为FSCIL鲁棒性研究的基石性论文

% 不同任务/数据集（如CIFAR-FSCIL, miniImageNet-FSCIL）下的攻击可迁移性，将增强结果的普适性。

\end{abstract}

% [可选] 添加关键词，ArXiv 支持显示
% \keywords{Few-Shot Learning \and Adversarial Attack \and Large Language Models \and Diffusion Models}

% --- 章节内容 ---
\section{Introduction}
\label{sec:intro}

\begin{figure}[t]
    \begin{subfigure}[b]{0.23\textwidth}
    \centering
    \includegraphics[width=0.98\textwidth]{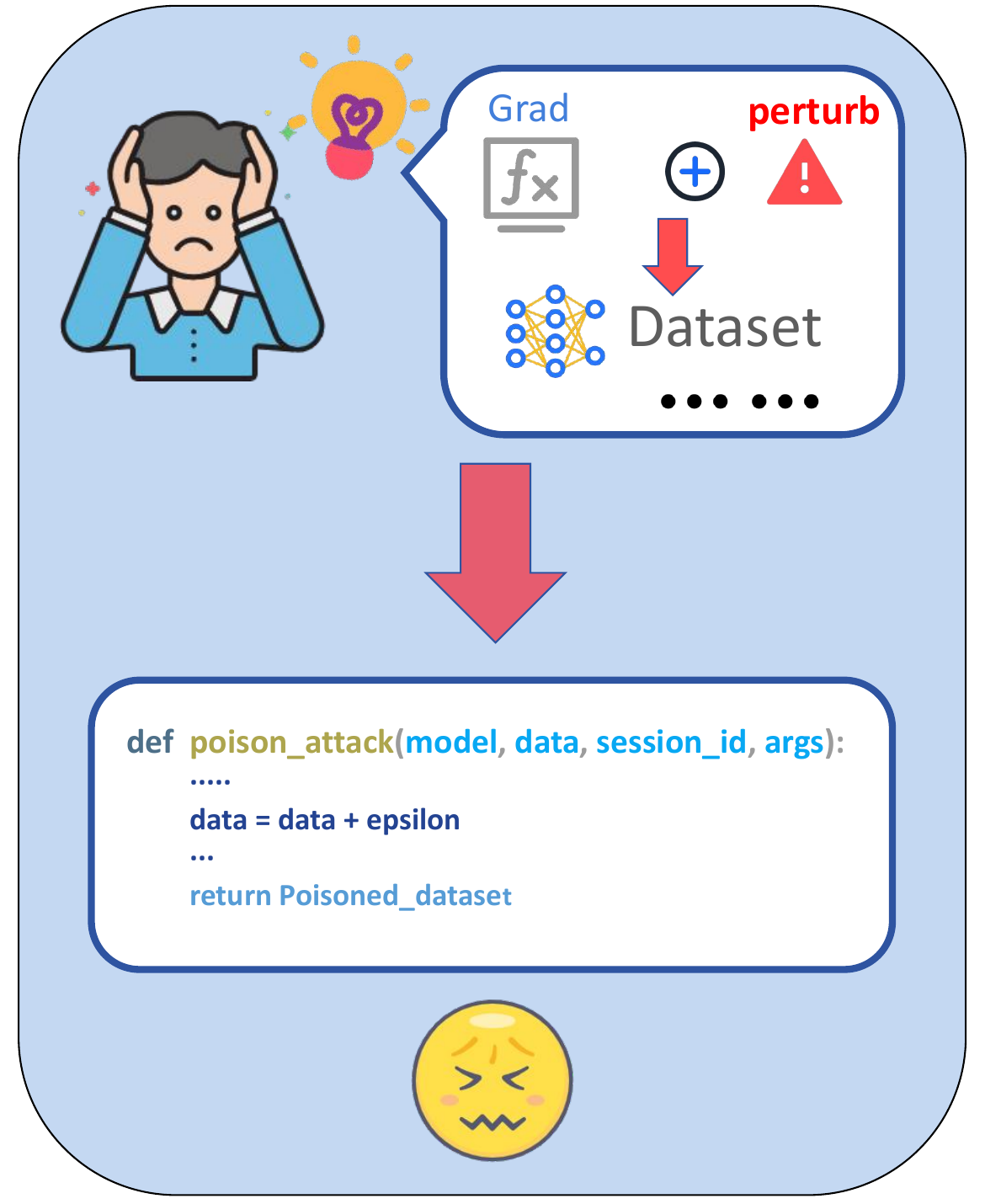}
    \caption{Manual Design}
    \label{fig:comparasion_a}
    \end{subfigure} \hfill
    \begin{subfigure}[b]{0.23\textwidth}
    \centering
    \includegraphics[width=0.98\textwidth]{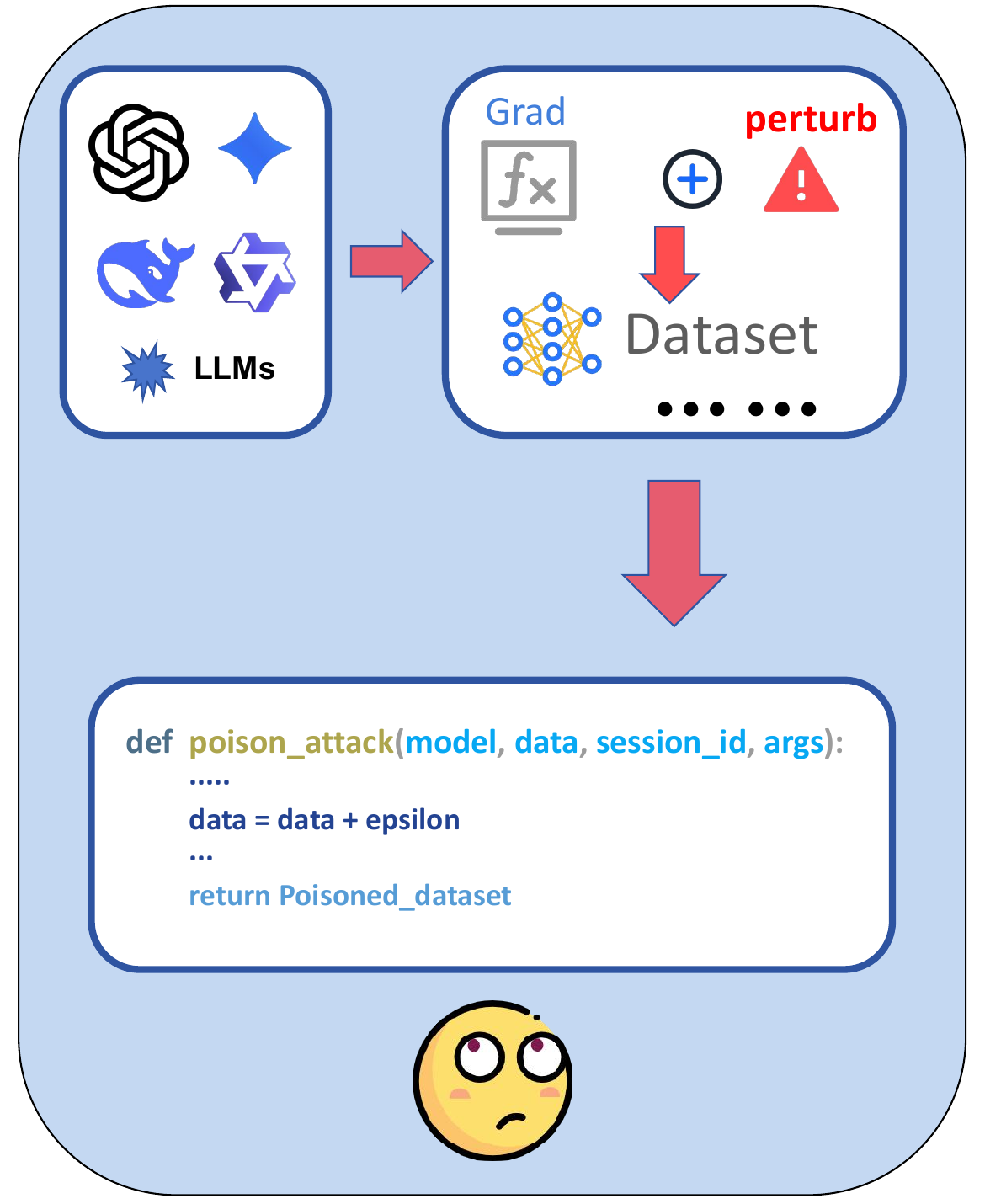}
    \caption{Naive method}
    \label{fig:comparasion_b}
    \end{subfigure}\hfill
    \centering
    \begin{subfigure}[b]{0.45\textwidth}
    \centering
    \includegraphics[width=0.98\textwidth]{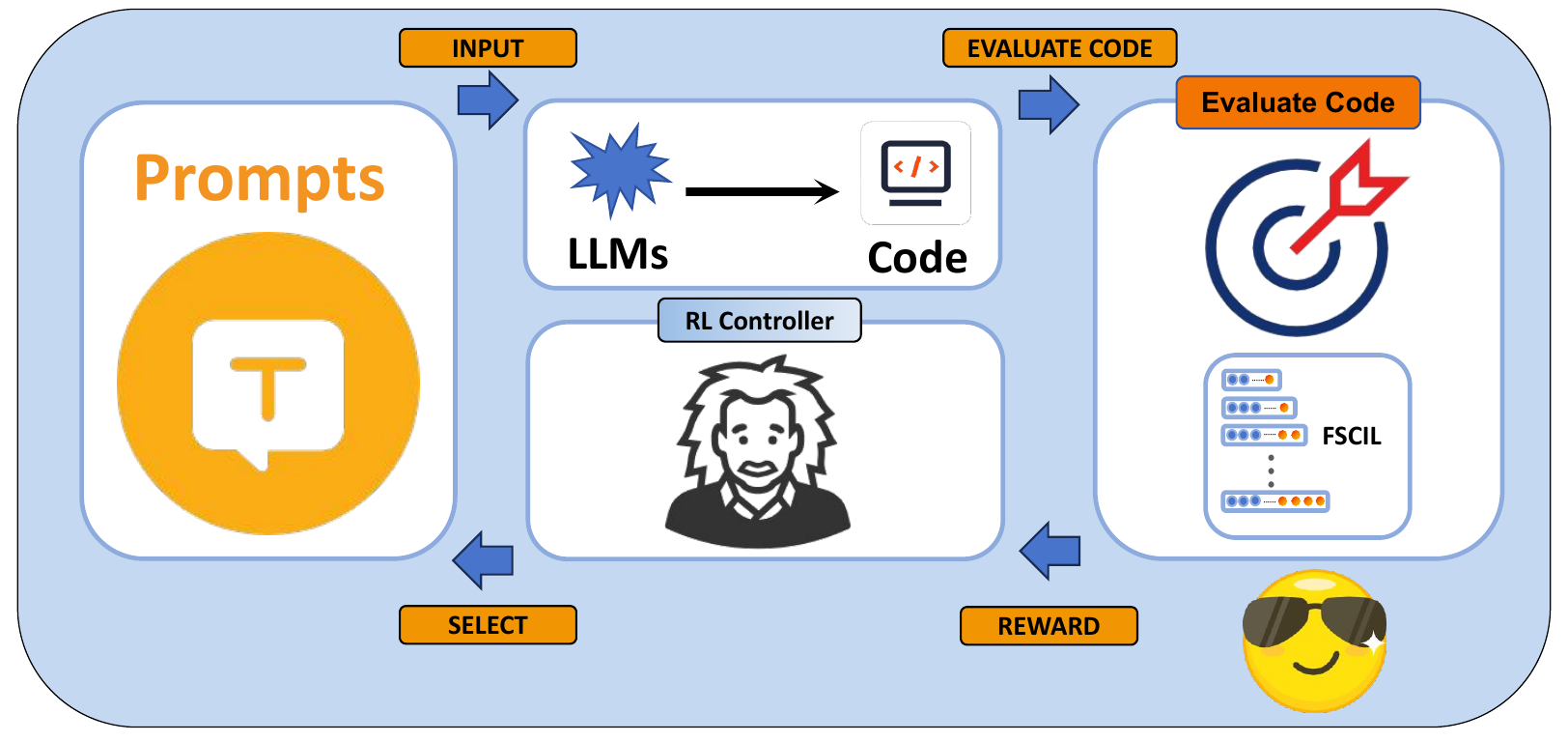}
    \caption{Ours}
    \label{fig:comparasion_c}
    \end{subfigure}
    \caption{ A comparison of the designed way of Attack. (a) Manual design relies on expert knowledge. (b) A naive method via LLMs. (c) Our method proposes an LLM-driven ACraft framework to automatically discover optimal Attack for FSCIL.} \vspace{-0.5cm} 
    \label{fig:comparasion}
\end{figure}

Deep neural networks have demonstrated remarkable success across various computer vision applications, showing superior performance in classification, detection, and segmentation tasks. However, this success typically relies on the assumption that the training and testing categories remain identical, and that models will not encounter new classes after deployment. This assumption often fails in real-world settings, where models must adapt to new categories over time, leading to catastrophic forgetting and degraded generalization. To mitigate this, Class-Incremental Learning (CIL) \cite{r17,r6} has emerged as a promising paradigm, enabling models to acquire new knowledge while retaining previously learned information.

Despite the progress of CIL, conventional paradigms still assume that each incremental task is supported by abundant labeled data, which is unrealistic in many applications such as robotic interaction, medical diagnosis updates, or recommendation systems. In such domains, new categories often emerge as few-shot instances, where only a handful of samples per class are available. This motivates the development of Few-Shot Class-Incremental Learning (FSCIL) \cite{r7,r8,r9,r10,r11}, which aims to bridge the gap between continual learning and data scarcity. Although recent advances have led to promising performance, the security vulnerabilities of FSCIL remain largely unexplored, posing significant threats in safety-critical applications.

\subsection{Challenges}

Although recent works have made initial attempts to explore adversarial attacks in the FSCIL setting, these methods still suffer from several fatal drawbacks:
(1) They require extensive expert knowledge through time-consuming manual design and tuning, which dramatically increases the cost and limits the scalability of attack algorithm development;
(2) They are usually tailored for traditional class-incremental learning, where the model remains relatively stable. In contrast, FSCIL models are highly non-stationary, as their decision boundaries continuously evolve with each incremental session. Consequently, perturbations optimized for one state of the model quickly become obsolete in subsequent states, leading to inconsistent attack efficacy;
(3) They fail to account for the inherent data sparsity in few-shot settings, resulting in biased or unstable attacks due to unreliable gradient estimation.

These limitations highlight the need to rethink the design paradigm of attack algorithms for FSCIL and explore new directions beyond the traditional handcrafted framework.

\noindent\textbf{Our New Observation.}
Different from existing manually designed methods (as shown in Fig. \ref{fig:comparasion_b} and Fig. \ref{fig:comparasion_c}), we observe a new way to automatically design adversarial attack algorithms via Large Language Models (LLMs) in this work. This paradigm fundamentally disrupts the traditional manual design process by allowing LLMs to autonomously generate and refine attack strategies, significantly reducing human effort while improving adaptability to dynamic FSCIL scenarios. More details are illustrated in Section \ref{sec:method}.

\subsection{Contributions}

In this work, we aim to analyze and overcome the aforementioned drawbacks by introducing a new paradigm of automated adversarial attack generation. To achieve this goal, we first explore a naive method that uses simple prompts as input to LLMs \cite{r18,r19,r20}. However, we find that this one-shot prompting paradigm performs poorly in FSCIL attack generation, as it degenerates into a black-box process with no interaction between the LLM and the target model. This observation raises a critical question: how can LLMs be effectively guided to generate more powerful and adaptive attack algorithms?

To address this challenge, inspired by actor-critic reinforcement learning and Chain-of-Thought (CoT) prompting \cite{CoT}, we propose a Proximal Policy Optimization (PPO)-based iterative framework that enables continuous refinement of LLM-generated attack algorithms. Specifically, the LLM first generates a population of diverse attack candidates, each of which is evaluated through a task-specific fitness function. A PPO-based controller then stabilizes policy updates through a clipping mechanism, producing improved prompts to guide the next generation of algorithm evolution. This closed-loop optimization fosters a positive feedback cycle between the LLM and the FSCIL task, leading to progressively stronger and more generalizable attack algorithms across iterations.

To ensure fairness and generality, we adopt a widely recognized FSCIL benchmark that supports training-time poisoning attacks, enabling comprehensive and standardized evaluation. Extensive experiments show that our proposed framework not only discovers novel and highly effective attack schemes but also outperforms expert-designed methods under the same computational constraints.

We summarize our main contributions as follows:

\begin{itemize}
\item \textbf{New Attack Generation Paradigm.} To the best of our knowledge, we are the first to propose a novel LLM-driven framework for automatically designing adversarial attacks in FSCIL, offering a fresh perspective for understanding and improving model robustness in continual few-shot scenarios.
\item \textbf{PPO-driven Evolutionary Strategy.} Beyond the limitations of simple prompting, we reveal that the absence of feedback signals is the root cause of poor attack quality. To overcome this, we introduce a PPO-based reinforcement learning mechanism that builds a reasoning-driven optimization loop, significantly enhancing the coherence and effectiveness of LLM-generated attacks.
\item \textbf{Numerical Verification.} Comprehensive experiments on standard FSCIL benchmarks demonstrate that our automated framework consistently outperforms manually designed attack methods, establishing a new state-of-the-art in automated adversarial algorithm discovery.
\end{itemize}

\begin{table*}[t]
\centering
\caption{\textbf{ACraft \textit{v.s.} previous methods.} $T$ denotes the total number of attack iterations and $d$ represents the input dimensionality. $Cost(\cdot)$ indicates the asymptotic computational complexity of each algorithm with respect to $T$ and $d$.}
\resizebox{\linewidth}{!}{
\begin{tabular}{cccccc}
\hline
\textbf{Method} & \textbf{Formula} & \textbf{Human Expert} & \textbf{Target}  & \textbf{Attack Acc$\downarrow$} & \textbf{Cost (Algorithm)} \\ \hline
FGSM \cite{goodfellow2014explaining} & $x_{adv} = \text{Clip}_{0,1} \left( x + \epsilon \cdot \text{sign} \left( \nabla_x \mathcal{L}(f(x), y) \right) \right)$  & \cmark & Image Classification & 58.01 & $O(d)$\\ 
 PGD \cite{madry2017towards} & $x^{t+1}_{adv} = \Pi_{x, \epsilon, [0,1]} \left( x^t_{adv} + \alpha \cdot \text{sign} \left( \nabla_x \mathcal{L}(f(x^t_{adv}), y) \right) \right)$  & \cmark & Image Classification & 58.00 & $O(T \cdot d)$ \\ 
C\&W \cite{carlini2017towards} & $\min_{w} \left\| \frac{1}{2}(\tanh(w) + 1) - x \right\|_2^2 + c \cdot f\left( \frac{1}{2}(\tanh(w) + 1) \right)$   & \cmark & Image Classification & 60.10 & $O(T \cdot d)$ \\ 
 DeepFool \cite{moosavi2016deepfool} & $\delta^{(t)} = \frac{\left| f_{k}(x^{(t)}) - \max_{j \neq k} f_j(x^{(t)}) \right|}{\left\| \nabla_x f_{k}(x^{(t)}) - \nabla_x f_{l}(x^{(t)}) \right\|_2^2} \left( \nabla_x f_{k}(x^{(t)}) - \nabla_x f_{l}(x^{(t)}) \right)$ & \cmark  & Image Classification  & 60.10 & $O(T \cdot d)$ \\ \hline
% \textbf{AutoPrune-v2} & $\left | W_{ij}\left ( \sqrt{Var\left [ X \right ] }+\sigma_{W }   \right )   \right | $ & \xmark   & \cmark & 128 & \xmark & $O(d^2)$ \\ 
\gr\textbf{ACraft} & $x_{\text{adv}} = \text{clip}\Bigg(
    \text{clip}\Bigg(
        x + \sum_{t=0}^{T-1} \Bigg[
            -\alpha \cdot \text{sign}\Bigg(
                \mu \cdot m^{(t)} + (1-\mu) \cdot \lambda_{\text{rev}} \cdot \nabla_x \mathcal{L}_{\text{comb}}^{(t)}
            \Bigg)
        \Bigg],
        x - \epsilon,
        x + \epsilon
    \Bigg),
    0,
    1
\Bigg)$ & \xmark & FSCIL & 17.13 & $O(T \cdot d)$ \\ 
\hline
\end{tabular}}
\label{tab:review_zc}
\vspace{-0.4cm}
\end{table*}

\begin{figure*}[t]
    \centering
    \includegraphics[width=\linewidth]{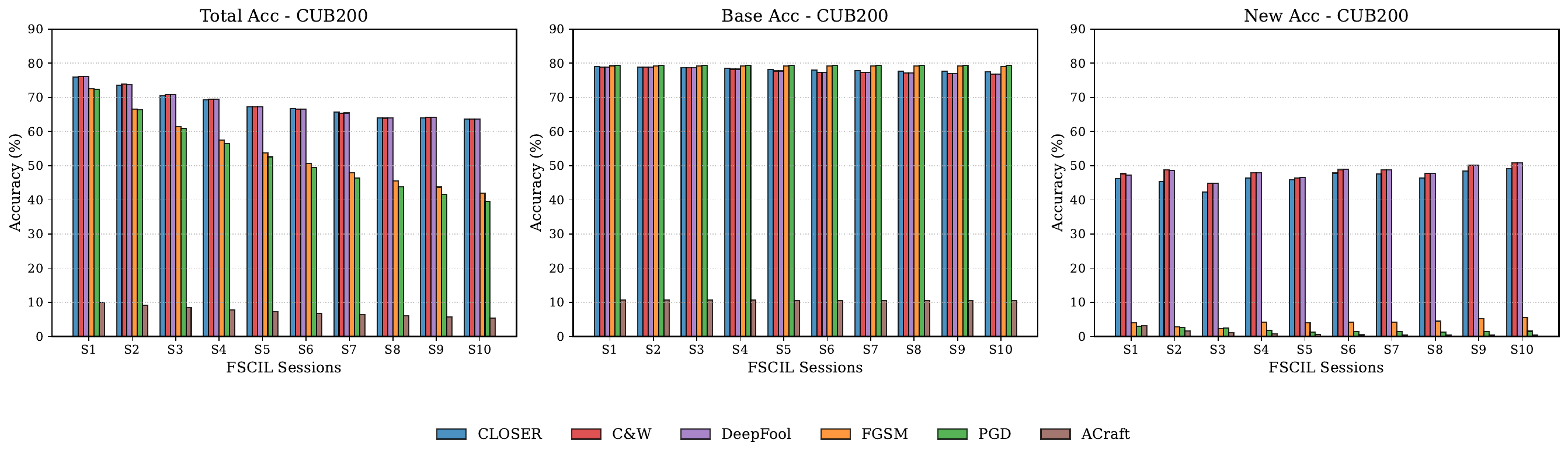}
    \caption{ACraft \textit{v.s.} existing attacks for FSCIL task in CUB200 dataset.} \vspace{-0.25cm}
    \label{fig:cub200}
\end{figure*}

\section{Rethinking the design of Attack Algorithms}
\label{RDZCP}
The primary objective of traditional attack algorithms is to efficiently attack general models. As shown in Table \ref{tab:review_zc}, representative attack algorithms (such as FGSM, PGD, and C\&W) can utilize heuristic or statistical methods to measure their expressiveness. However, these methods heavily rely on specialized expert knowledge, making them suboptimal when applied to specific models. Additionally, the design of such algorithms is particularly time-consuming. As illustrated in Table \ref{tab:review_zc}, the attack algorithms manually designed by experts exhibit extremely low effectiveness when used for FSCIL attacks. Moreover, to validate the statement that existing attack methods either fail to attack base classes, we conduct an in-depth analysis in the CUB-200 dataset tailored for the FSCIL task. As depicted in Fig. \ref{fig:cub200}, we can clearly observe that existing attack methods (e.g., PGD, DeepFool, etc.) indeed do not work on base classes. In addition, ``C\&W" and ``DeepFool" fail to attack on new classes.  By contrast, our Arcraft significantly attacks the base and new classes, leading to an approximate
 70\%, and 45\% decrease in terms of accuracy, respectively.

Grounding in those findings, we analyze that those issues may be related to the task of FSCIL, where the small perturbations independently optimized for a few samples attacked by methods (e.g., C\&W) are offset during the incremental update of the FSCIL model, resulting in a poisoning prototype that is almost identical to the clean prototype and unable to generate destructive parameter updates. The low performance of these manually crafted adversarial attack algorithms for FSCIL prompted us to seek a new paradigm for designing attack methods specifically tailored for FSCIL.

Inspired by the success of LLMs in generating new knowledge, we pose a natural question: \textbf{Can LLMs automatically generate adversarial attack algorithms for FSCIL?} To rigorously answer this question, we first designed a Naive method (as shown in \textbf{\red{App. A}}), asking the LLM to output a reasonable attack idea along with the corresponding implementation code. To validate the effectiveness proposed naive method, we conduct an in-depth analysis based on CLOSER~\cite{oh2024closer} in the mainstream miniImageNet dataset. As shown in the Table. \ref{tab:LLM}, the generated code reduced model accuracy by an average of 19.10\% under four runs, which still remains lower than manually designed methods (i.e., PGD: 4.76\% \textit{v.s.} Naive: 19.10\%). Although the results are promising, the Naive method is still far from the expectation of attacking for FSCIL. Further analysis demonstrates that the limitation of the Naive method arises from the absence of feedback from FSCIL: attack code generation conditioned only on prompts forms a blackbox process. To address this limitation, we propose an iterative optimization algorithm based on PPO reinforcement learning. Our framework incorporates a feedback mechanism that enables closed-loop optimization, allowing the LLM to iteratively refine the generated attack algorithms. This significantly improves the performance of ACraft (as shown in Fig. \ref{fig:cub200}).
\section{Automatic Attack Discovery for FSCIL}
\label{sec:method}

\subsection{Overview of ACraft}
To generate novel attack algorithms for FSCIL, ACraft leverages LLMs as a powerful tool to produce new attack strategies along with their corresponding code. However, LLMs can sometimes behave like black boxes. To address this issue, ACraft integrates the PPO reinforcement learning algorithm, which provides continuous feedback on the effectiveness of generated attacks and uses this feedback to guide the LLM’s generation process intelligently. The ACraft system consists of three cooperating components: the Adversarial Algorithm Generator, the Attack Efficacy Evaluator, and a PPO-based Evolution Controller.

\noindent \textbf{Adversarial Algorithm Generator.} In the ACraft framework, the Large Language Model (LLM) is tasked with generating novel attack algorithms, guided by a series of carefully designed prompts. These prompts instruct the LLM to either refine an existing attack method or synthesize features from multiple methods to create a more effective one. Unlike typical code generators, each LLM output consists of two interconnected components for every attack algorithm. First, it produces a Strategic Rationale in natural language, explaining the core logic of the attack. Second, it generates the executable Algorithmic Code. This two-part output is crucial because it ensures that the generated method is interpretable and allows us to clearly understand how and why each attack works.
\begin{figure*}[t]
    \centering
    \includegraphics[width=\linewidth]{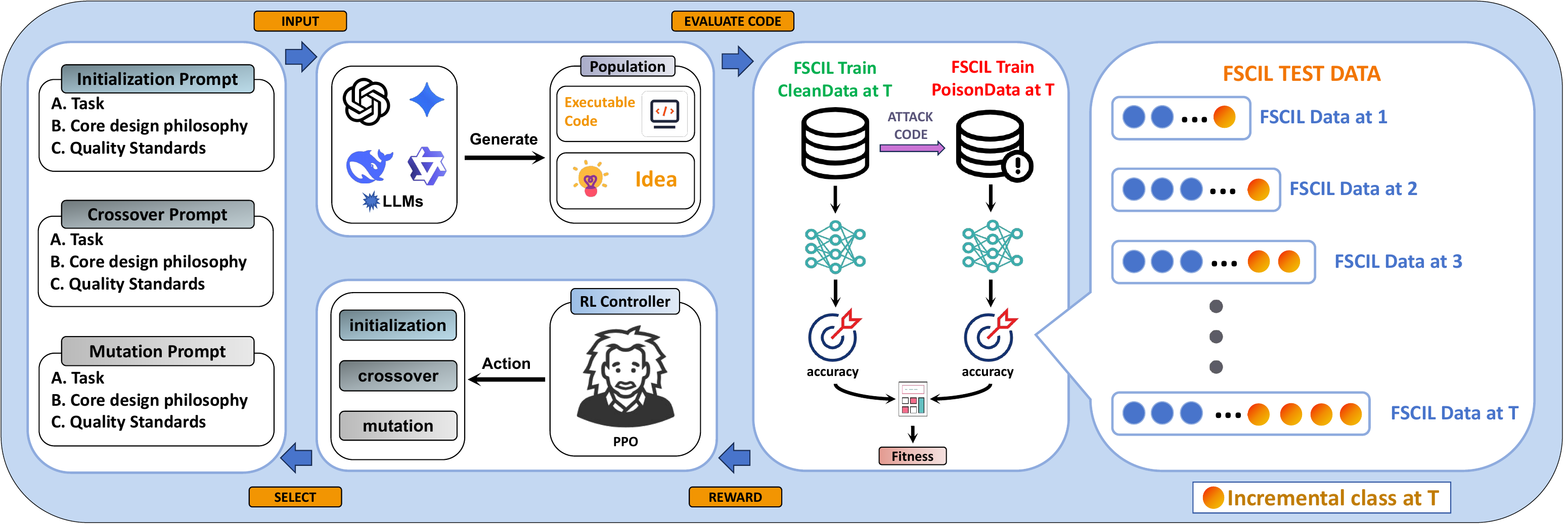}
    \caption{Overview of ACraft.} \vspace{-0.5cm}
    \label{fig:comparison}
\end{figure*}

\noindent \textbf{Attack Efficacy Evaluator.} The primary task of this evaluator is to quantify the effectiveness of each generated attack algorithm. To this end, it performs attacks on a target FSCIL model and measures their impact on the model's performance. These measurements are then aggregated into a single fitness score, a comprehensive metric that primarily accounts for both the extent of damage caused to the model's performance and the computational cost of the attack. Ultimately, this fitness score serves as a crucial feedback signal to guide the entire optimization process.

\noindent \textbf{PPO-based Evolution Controller.} The PPO controller is the decision-maker for the whole evolution process. Its primary task is to learn the optimal policy to guide the generation process. In this way, the evolution can find the best attack algorithm. To this end, the controller uses the fitness score from the evaluator as a reward signal. It then employs the PPO algorithm to learn the best strategy. This strategy helps it decide the next action: whether to instruct the generator to create a new algorithm or to fine-tune an existing algorithm that shows promise.

\subsection{Mathematical Formulation}
We consider a standard FSCIL scenario. In this setting, a model \(\mathcal{M}\) is trained on a series of incremental tasks \(\mathcal{T}=\{\mathcal{T}_1, \mathcal{T}_2, ..., \mathcal{T}_N\}\) that arrive in order. Our goal is to design a general attack method. This method works by adding poisoned samples to the training data \(\mathcal{D}_{train}^{\mathcal{T}_i}\) of each task \(\mathcal{T}_i\). In this way, it continuously damages the model's performance. We structure the collection of all possible attack algorithms as a latent design space \(\mathcal{G}\) encoded by LLM. Each attack algorithm \(\in \mathcal{G}\) from this space is represented as a set of two-components. The first part of this set is a conceptual description \(R\) of the attack algorithm that the LLM designs. It explains the logic of the attack, and then LLM translates this conceptual description into a fully executable implementation \(C\). This implementation forms the second part of the set.

To measure the effectiveness of any algorithm \(g\), we define an objective function \(\mathcal{L}_{attack}: \mathcal{M} \times \mathcal{D} \rightarrow \mathbb{R}\). We use \(P(M_\theta, \mathcal{D})\) to show how well a model with parameters \(\theta\) performs on a dataset \(\mathcal{D}\). For example, this performance can be the accuracy. Next, we define two models. We let \(M_{clean}\) be the model that trains on the original, clean data. In contrast, \(M_g\) is the model that trains on the poisoned data. Therefore, the objective function \(\mathcal{L}_{attack}\) is the performance difference between these two models:
\[\mathcal{L}_{attack}(M_g, \mathcal{D}_{test}) = P(M_{clean}, \mathcal{D}_{test}) - P(M_g, \mathcal{D}_{test})\]
Formally, let \(M_g\) represent the model after the original model \(M\) undergoes poisoned training, as specified by \(C\), the implementation of algorithm \(g\). This implementation specifies the poisoned training procedure. The final goal of the ACraft framework is to find the best attack algorithm \(g^*\). This algorithm aims to maximize the expected effectiveness of the attack on the distribution of the test data. This optimization problem can be represented as:
\[g^* = \arg\max_{g \in \mathcal{G}} \mathbb{E}_{\mathcal{D}_{test}} [\mathcal{L}_{attack}(M_g, \mathcal{D}_{test})]\]
Here, the expectation \(\mathbb{E}\) shows that we are looking for an attack strategy that can generalize well. The entire ACraft framework is designed specifically to solve this optimization problem effectively.

\subsection{Core Mechanisms of Attack Generation}

\noindent \textbf{Algorithm Generator.}
We model the Algorithm Generator as a set of discrete transformation functions \(\mathcal{F}_{trans}\). These functions make up the executable operation space used by the generator when creating attack algorithms. Theoretically, these functions are specific, controllable mappings that the generator performs on its internally encoded latent design space \(\mathcal{G}\).

\noindent First, we define \(\Pi\) as the set of all possible instructions. Each function \(F \in \mathcal{F}_{trans}\) has a fixed structure. Its main role is to turn the input context \(\mathcal{C}\) into executable code \(g'\):
\[F: \mathcal{C} \rightarrow \mathcal{G}\]
where \(\mathcal{C}\) is a tuple made of all possible input information for the generator. It consists of an instruction subset \(\Pi_{sub} \subseteq \Pi\) and a parent algorithm subset \(\mathcal{G}_{sub} \subseteq \mathcal{G}\):
\[\mathcal{C} = (\Pi_{sub}, \mathcal{G}_{sub})\]
The set of transformation functions \(\mathcal{F}_{trans}\) includes the following types:

\begin{itemize}

\item \noindent \textbf{Genesis Function \(F_{init}\):} 
This function creates entirely new algorithms from scratch. In this case, the parent set \(\mathcal{G}_{sub}\) is empty \(\emptyset\). Therefore, the input context \(\mathcal{C}\) relies only on the initial instruction \(\Pi_{init} \in \Pi_{sub}\).
\[g' = F_{init}(\Pi_{init}, \emptyset)\]
When the Algorithm Generator runs this function, its goal is to maximize the diversity measure \(D(\mathcal{G}_{new})\) of the newly created subset \(\mathcal{G}_{new}\). This ensures that the search is not limited to a narrow part of \(\mathcal{G}\).

\item \noindent \textbf{Refinement Function \(F_{refine}\):}
This function adjusts an existing successful algorithm \(g \in \mathcal{G}\) using fine-tuning and local optimization (exploitation). This helps the Generator reach higher fitness peaks in the strategy space. Its input context \(\mathcal{C}\) includes a single parent algorithm \(\{g\} \subseteq \mathcal{G}\) and a specific optimization instruction \(\Pi_{opt} \in \Pi_{sub}\).
\[g' = F_{refine}(\Pi_{opt}, \{g\})\]
\noindent When the Generator runs this function, it must ensure that the new executable code \(g'\) maintains a high correlation with \(g\) in its strategy: \(Corr(g', g) \approx 1\) while simultaneously achieving a performance improvement: \(\phi(g') > \phi(g)\).

\item \noindent \textbf{Synthesis Function \(F_{synth}\):} This function combines the best strategies from multiple parent algorithms. It encourages strategy mixing and breakthrough innovation (exploration). Its input context \(\mathcal{C}\) includes a set of parent algorithms \(\{g_1, g_2, ..., g_k\} \subseteq \mathcal{G}\) and a merging instruction \(\Pi_{merge} \in \Pi_{sub}\).

\[g' = F_{synth}(\Pi_{merge}, \{g_1, g_2, ..., g_k\})\]
\end{itemize}

\noindent The main goal when the Generator runs this function is to maximize the expected effectiveness of the new executable code: \(\mathbb{E}[\phi(g')]\). This particularly targets the nonlinear gain that results from combining \(k\) different strategies.

\noindent By making the LLM's generation capability into this discrete and controllable function set \(\mathcal{F}_{trans}\), we successfully change the traditional problem of black-box code generation into a decision problem that a reinforcement learning strategy can clearly select and manage. This function-based modeling not only makes the theoretical analysis and tuning of the algorithm generation process possible, but also ensures that the ACraft framework can perform a systematic, intelligent evolutionary search among diversity injection \(F_{init}\), local optimization \(F_{refine}\), and strategy breakthrough \(F_{synth}\).

\noindent \textbf{Fitness Evaluator as a Multi-Attribute Utility Function.}
The primary job of the Evaluator is to calculate a fitness score \(\phi(g)\) for each algorithm \(g\). This score is modeled as a Multi-Attribute Utility Function, balancing attack effectiveness and cost to reflect the algorithm's real-world value.
First, we define an objective vector \(\vec{J}(g)\). This vector captures all performance metrics that are important to us:
\[\vec{J}(g) = [J_{succ}(g), J_{cost}(g)]^T\]
where \(J_{succ}(g)\) is further defined as a weighted sum of performance drops in new and old tasks. This shows how well the attack strategy targets catastrophic forgetting:
\begin{align*}
J_{succ}(g) &= \alpha \cdot \left( P(M_{clean}, \mathcal{D}_{old}) - P(M_g, \mathcal{D}_{old}) \right) \notag \\
&+ (1-\alpha) \cdot \left( P(M_{clean}, \mathcal{D}_{new}) - P(M_g, \mathcal{D}_{new}) \right)
\end{align*}
where \(J_{cost}(g)\) stands for the comprehensive cost, which includes the computational expense and the perturbation budget (constraining stealthiness). \(\alpha \in [0, 1]\) is a hyperparameter used to adjust how much we prioritize the forgetting of old knowledge. To obtain a single fitness score \(\phi(g)\), we perform a linear scalarization of the objective vector. We use a preset weight vector \(\vec{w} = [w_{succ}, w_{cost}]^T\), where \(w_{cost}\) is a negative value. The fitness is then calculated as: \[\phi(g) = \vec{w} \cdot \vec{J}(g) = w_{succ}J_{succ}(g) + w_{cost}J_{cost}(g)\]
The weight vector \(\vec{w}\) encodes our preference among objectives, ensuring that the search for the optimal algorithm \(g^*\) balances high attack effectiveness with low execution cost.

\noindent \textbf{The Controller: Optimal Policy Learning via PPO.}
The Controller is the decision core of the ACraft. Its task is to learn an optimal evolutionary strategy \(\pi_{\theta}\). This strategy aims to maximize the expected fitness of the new algorithms generated by the set of transformation functions \(\mathcal{F}_{trans}\). We model this strategy learning process as a policy optimization problem, and we use the PPO algorithm to solve it.

\noindent The parameters \(\theta\) of the strategy \(\pi_{\theta}\) define the probability distribution \(\pi_{\theta}(a_t|s_t)\) that the Controller uses to select an evolutionary action \(a_t\) given the population state \(s_t\). A core benefit of PPO is its ability to update the policy stably, preventing performance collapse caused by excessively large policy steps.
\begin{itemize}
    \item State (\(s_t\)): \(s_t\) describes the statistical properties of the current algorithm population \(P_t\) and the history of actions.
    \item Action (\(a_t\)): The action space \(\mathcal{A}\) is the set of decisions to select a transformation function \(F \in \mathcal{F}_{trans}\) and choose its required context \(\mathcal{C}\). Thus, \(a_t = (F, \mathcal{C})\).
\end{itemize}

\begin{algorithm}[t]
\caption{The ACraft Evolutionary Loop}
\label{alg:ACraft}
\begin{algorithmic}[1]
\Require LLM Generator \(\mathcal{L}\), Benchmark \(\mathcal{B}\), Population Size \(N\), Budget \(T_{max}\), PPO Controller \((\pi_{\theta}, V_{\psi})\)
\State Define initial instruction set \(\Pi_{init}\) based on \(\mathcal{B}\)
\State \(\mathcal{P}_{0} \leftarrow F_{init}(\Pi_{init}, \emptyset)\) 
\State \(s_{0} \leftarrow [\mu_{\Phi_0}, \sigma^2_{\Phi_0}, \emptyset]\)
\For{\(t=1\) to \(T_{max}\)}
    \State \(a_{t} \leftarrow \pi_{\theta}(\cdot|s_{t})\) 
    \State \(G'_{t} \leftarrow F(\mathcal{C})\) 
    \For{each \(g' \in G'_{t}\)}
        \State Compute fitness \(\phi(g') \leftarrow \vec{w} \cdot \vec{J}(g')\)
    \EndFor
    \State \(r_{t} \leftarrow \frac{1}{|G'_{t}|} \sum_{g' \in G'_{t}} \phi(g')\) 
    \State Update Controller \((\pi_{\theta}, V_{\psi})\) using \(L^{CLIP}(\theta)\) and \(r_{t}\) 
    \State \(s_{t+1} \leftarrow [\mu_{\Phi_{t+1}}, \sigma^2_{\Phi_{t+1}}, \text{History}(a_{t-k:t-1})]\) 
    \State \(\mathcal{P}_{t+1} \leftarrow \text{SelectTopN}(\mathcal{P}_{t} \cup G'_{t})\) 
    \If{average fitness has converged}
        \State \textbf{break}
    \EndIf
\EndFor
\State \Return Best algorithm found, \(g^*\) 
\end{algorithmic}
\end{algorithm}

\noindent To achieve the robust convergence of the strategy, PPO optimizes a clipped objective function \(L^{CLIP}(\theta)\). This function limits how much the new policy \(\pi_{\theta}\) can change relative to the old policy \(\pi_{\theta_{old}}\). 
\[L^{CLIP}(\theta) = \hat{\mathbb{E}}_t \left[ \min(r_t(\theta)\hat{A}_t, \text{clip}(r_t(\theta), 1-\epsilon, 1+\epsilon)\hat{A}_t) \right]\]
The core components of the PPO objective are defined as follows:
\begin{itemize}
\item Policy Ratio (\(r_t(\theta)\)): This ratio measures the probability of the new policy, \(\pi_{\theta}\), selecting the action \(a_t\) relative to the old policy, \(\pi_{\theta_{old}}\).
\[r_t(\theta) = \frac{\pi_{\theta}(a_t|s_t)}{\pi_{\theta_{old}}(a_t|s_t)}\]

\item Reward (\(r_t\)): This represents the immediate gain signal for the PPO optimization. It is defined as the average fitness score of the new generation \(G'_{t}\) produced after executing action \(a_t\):\[\text{Reward } r_t = \frac{1}{|G'_{t}|} \sum_{g' \in G'_{t}} \phi(g')\]
\end{itemize}
where:
\begin{itemize}
\item \(\phi(g')\) is the fitness score of the generated algorithm \(g'\), which balances attack efficacy and cost (as defined by the Multi-Attribute Utility Function).
\item \(\hat{A}_t\) is the Advantage Estimate, which shows the expected extra return of the current action \(a_t\) compared to the average policy.
\item The Clipping Term limits the policy ratio \(r_t(\theta)\) to the range \([1-\epsilon, 1+\epsilon]\), where \(\epsilon\) is a clipping hyperparameter in PPO.
\end{itemize}
By iteratively optimizing \(L^{CLIP}(\theta)\), the Controller learns a meta-level strategy \(\pi_{\theta}\) efficiently. This strategy can intelligently orchestrate the algorithm generation process (that is, select the optimal transformation function \(F \in \mathcal{F}_{trans}\)) to systematically improve the overall fitness of the population. Consequently, it efficiently solves the optimization problem defined in the Mathematical Formulation section: \(g^* = \arg\max_{g \in \mathcal{G}} \mathbb{E}[\mathcal{L}_{attack}(M_g)]\).
\subsection{The Closed-Loop Algorithmic Workflow}

This section integrates the Generator, Evaluator, and Controller into an iterative evolutionary cycle. This closed-loop workflow is essential for ACraft to efficiently solve for the optimal strategy \(g^*\). The cycle systematically explores the latent design space \(\mathcal{G}\) by continuously feeding algorithm performance back into the strategy optimizer.

\begin{table*}[t]
\definecolor{colorfirst}{RGB}{255, 153, 154} 
\definecolor{colorsecond}{RGB}{255, 255, 151} 
\definecolor{colorthird}{RGB}{255, 204, 153}
\caption{\textbf{Performance comparison in miniImageNet dataset.} Session-wise Acc (\%) and the average of them (Avg) are reported. Attack Drop. represent the magnitude of the decrease in the accuracy of the last session after being attacked.}
%\vspace{-0.15in}
\label{table:comp_mini}
\centering
\resizebox{\linewidth}{!}
{
    \begin{tabular}{lcccccccccccr}
    \toprule
    
    \multirow{2}{*}{\textbf{Methods}}\; & \multirow{2}{*}{\textbf{Year}}\; &\multicolumn{9}{c}{\textbf{Acc in each session} (\%)$\downarrow$} & \multirow{2}{*}{\textbf{Avg$\downarrow$}} \;& \multirow{2}{*}{\makecell{\textbf{Attack.} \\ \textbf{Drop.$\uparrow$}}}\\
    \cmidrule{3-11}
    & \;& 0  \;& 1  \;&2  \;& 3  \;& 4  \;& 5  \;& 6  \;& 7  \;& 8  \;&    \;& \\
    \midrule
    CLOSER~\cite{oh2024closer} & ECCV 2024\;& 76.02 \;&71.61 \;&67.99 \;&64.69 \;&61.70 \;&58.94 \;&56.23 \;&54.52 \;&53.33 \;&62.78 \;& -\\
    \midrule
    FGSM \cite{goodfellow2014explaining} & \;& 76.02 \;&\cellcolor{colorsecond}68.15 \;&\cellcolor{colorthird}63.73 \;&\cellcolor{colorsecond}59.98 \;&\cellcolor{colorsecond}56.46 \;&\cellcolor{colorthird}53.33 \;&\cellcolor{colorthird}50.47 \;&\cellcolor{colorsecond}47.93 \;&\cellcolor{colorthird}46.08 \;&\cellcolor{colorthird}58.01\;& \cellcolor{colorsecond}4.77\\
    PGD \cite{madry2017towards} & \;& 76.02 \;&\cellcolor{colorthird}68.4 \;&\cellcolor{colorsecond}63.63 \;&\cellcolor{colorthird}60.09 \;&\cellcolor{colorthird}56.53 \;&\cellcolor{colorsecond}53.12 \;&\cellcolor{colorsecond}50.36 \;&\cellcolor{colorthird}47.98 \;&\cellcolor{colorsecond}45.87 \;&\cellcolor{colorsecond}58.00\;& \cellcolor{colorthird}4.76\\
    C\&W \cite{carlini2017towards} & \;& 76.02\;&68.98 \;&64.94 \;&61.73 \;&58.79 \;&55.93 \;&53.27 \;&51.16 \;&50.13 \;&60.10\;& 1.68\\
    DeepFool \cite{moosavi2016deepfool} & \;& 76.02 \;&68.98 \;&64.94 \;&61.73 \;&58.79 \;&55.93 \;&53.27 \;&51.16\;&50.13 \;&60.10\;& 1.68\\
    \midrule
    \textbf{ACraft (our)} & \;& 76.02 \;&\cellcolor{colorfirst}{12.20} \;&\cellcolor{colorfirst}{11.34} \;&\cellcolor{colorfirst}{10.52} \;&\cellcolor{colorfirst}{9.88} \;&\cellcolor{colorfirst}{9.31} \;&\cellcolor{colorfirst}{8.69} \;&\cellcolor{colorfirst}{8.28} \;&\cellcolor{colorfirst}{7.96} \;&\cellcolor{colorfirst}{17.13}\;& \cellcolor{colorfirst}{58.89}\\

    \bottomrule
    \end{tabular}
}
\end{table*}

\begin{figure*}[t]
    \centering
    \includegraphics[width=\linewidth]{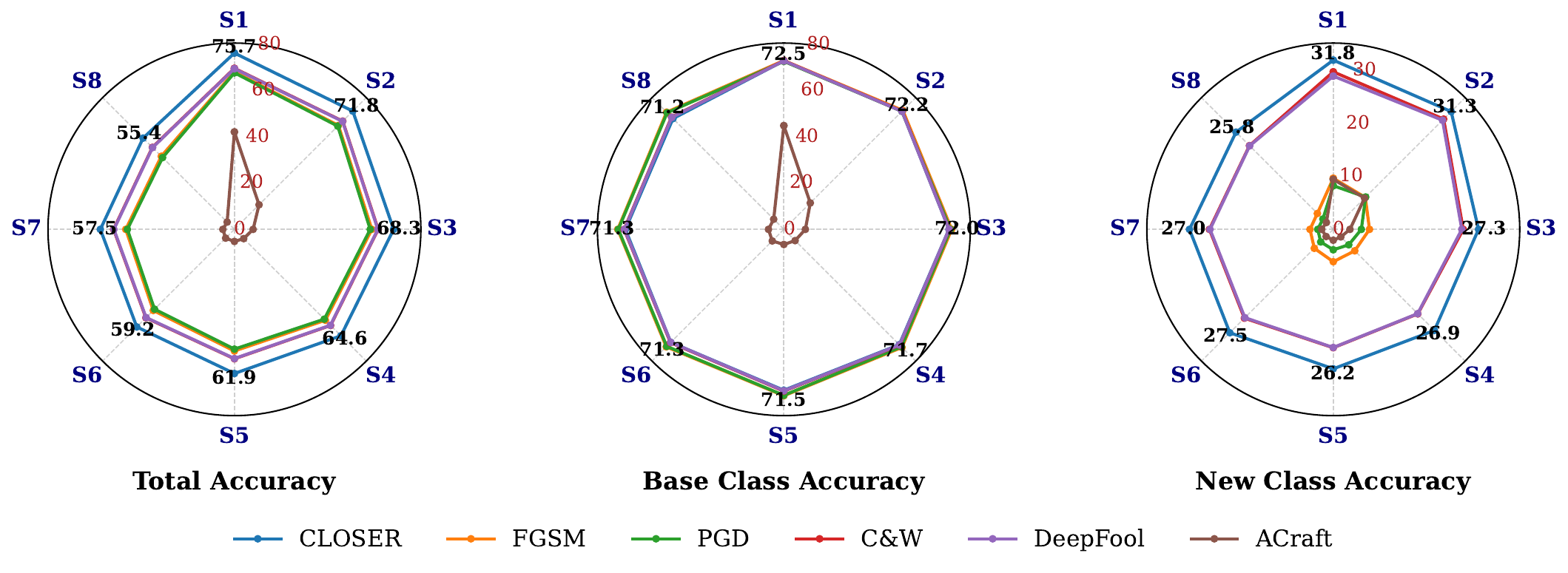}
    \caption{ACraft \textit{v.s.} existing attacks for FSCIL task in Cifar-100 dataset.} \vspace{-0.25cm}
    \label{fig:cifar10}
\end{figure*}

\noindent \textbf{Step 0: Initialization and Seed Generation.} The Controller selects the Genesis Function \(F_{init}\), and the Generator executes \(g' = F_{init}(\Pi_{init}, \emptyset)\) until \(N\) initial algorithms form the population \(P_0\).

\noindent \textbf{Step 1: Policy Sampling and Algorithm Synthesis.} The PPO Controller observes the current state \(s_t\), samples action \(a_t = (F, \mathcal{C})\) based on \(\pi_{\theta}\), and the Generator executes \(g' = F(\mathcal{C})\) to create the offspring \(G'_t\).

\noindent \textbf{Step 2: Attack Deployment and Fitness Evaluation.} Each \(g' \in G'_t\) is deployed against the target system. The Evaluator runs the executable code \(C\), measures the impact, and calculates the fitness \(\phi(g')\) using the Multi-Attribute Utility Function. Failed algorithms receive a penalty value.

\noindent \textbf{Step 3: Strategy Update via PPO.} The average fitness \(r_t = \mathbb{E}[\phi(G'_t)]\) is fed back as the reward. PPO uses this reward and the policy ratio \(r_t(\theta)\) to optimize the clipped objective function \(L^{CLIP}(\theta)\), updating \(\pi_{\theta}\) parameters \(\theta\).

\noindent \textbf{Step 4: Population Management and Iteration.} The candidate set \(P_t \cup G'_t\) is ranked by \(\phi(\cdot)\). The top \(N\) algorithms are kept to form \(P_{t+1}\). The loop continues until the maximum iterations \(T_{max}\) are reached or the average fitness \(\mu_{\Phi}\) converges. The final output is the best discovered algorithm is \(g^* = \max_{g \in \mathcal{G}} \phi(g)\).

\section{Experiment and Discussions}
\subsection{Experimental Settings}
We conduct extensive experimental comparison on benchmarks (i.e., miniImageNet~\cite{matchingnet}, CIFAR100~\cite{krizhevsky2009learning}, and CUB200~\cite{wah2011caltech}) to validate the effectiveness of the proposed method. We take CLOSER~\cite{oh2024closer} as our main baseline. The detailed experimental settings are presented in \textbf{\red{App. B}}.

\begin{table*}[t]
\definecolor{colorfirst}{RGB}{255, 153, 154} 
\definecolor{colorsecond}{RGB}{255, 255, 151} 
\definecolor{colorthird}{RGB}{255, 204, 153}
\caption{\textbf{Generalizability of our ACraft for more FSCIL methods in miniImageNet dataset.}}
\label{table:head_ablation}
%\vspace{-0.2cm}
\centering
\renewcommand{\arraystretch}{1.0}
\resizebox{0.9\linewidth}{!}
{
    \begin{tabular}{lcccccccccccc}
    \toprule
    \multirow{2}{*}{\textbf{FSCIL methods}} &\multirow{2}{*}{\textbf{Year}} & \multicolumn{9}{c}{\textbf{Acc in each session} (\%)$\downarrow$} && \multirow{2}{*}{\textbf{Avg}$\downarrow$}\\
    
    \cmidrule{3-11}
    && 0 & 1 & 2 & 3 & 4 & 5 & 6 & 7 & 8 && \\
    %&& & 0 & 1 & 2 & 3 & 4 & 5 & 6 & 7 & 8 &  \\
    
    % && $\phi_0$ && $\phi_{\text{old}}$ && $\phi_{\text{all}}$ &&  & & & & & & & &\\
    \midrule
    Limit \cite{zhou2023few} &TPAMI2023&84.13\;&78.80\;&74.21\;&69.64\;&66.01\;&63.32\;&61.50\;&58.17\;&55.88\;&&67.96\\
    % \midrule
    Limit + ACraft& &84.13\;&18.45\;&16.89\;&15.28\;&14.29\;&13.12\;&12.35\;&11.63\;&10.88\;&&19.65 $\downarrow$ (48.31)\\\midrule
    Approximation \cite{wang2024approximation} &ECCV2024& 
    84.13\;&75.25\;&70.25\;&67.54\;&64.85\;&61.34\;&59.74\;&56.38\;&53.45\;&&65.88\\
    Approximation + ACraft && 84.13\;&17.82\;&15.61\;&14.08\;&12.97\;&11.34\;&10.78\;&9.92\;&9.45\;&&18.01 $\downarrow$ (47.87)\\\midrule
    OrCo~\cite{ahmed2024orco}  &CVPR2024&84.13\;&77.84\;&73.42\;&69.22\;&65.85\;&62.75\;&60.56\;&57.34\;&55.33\;&&67.38\\
    OrCo + ACraft  &&84.13\;&19.06\;&16.84\;&15.30\;&13.47\;&12.48\;&11.50\;&10.61\;&9.80\;&&19.13 $\downarrow$ (48.07)\\\midrule
    Tri-WE \cite{zhou2023few} &CVPR2025&84.13\;&81.41\;&76.65\;&73.59\;&70.1\;&65.13\;&63.42\;&61.02\;&60.13\;&&70.62\\
    % \midrule
    Tri-WE \cite{zhou2023few} + ACraft&&84.13\;&20.35\;&17.99\;&15.76\;&14.09\;&12.62\;&11.08\;&10.92\;&10.12\;&&19.56 $\downarrow$ (51.06)\\
    
    \bottomrule
    \end{tabular}
}
\end{table*}

\begin{table}[t]
\centering
\caption{Comparison of ``Attack Drop'' performance between Naive and ACraft methods with different LLMs on CLOSER~\cite{oh2024closer} in the miniImageNet dataset. \label{tab:LLM}}
\renewcommand{\arraystretch}{1.2}
\resizebox{0.65\linewidth}{!}{%
\begin{tabular}{lcccccr}
\toprule
\textbf{Methods} & \textbf{LLMs}           & \textbf{Run 1} & \textbf{Run 2} & \textbf{Run 3} & \textbf{Run 4} & \textbf{Average} \\
\midrule
Naive     & gpt-4o-mini \cite{achiam2023gpt}  & 19.37  & 18.74  & 18.52 & 19.76 & 19.10  \\\midrule
ACraft    & Claude 3.7 \cite{TheC3} & 53.76  & 55.28  & 52.72 & 48.93 & 52.67        \\
ACraft    & Deepseek V3 \cite{liu2024deepseek}  & 49.32  & 45.77  & 50.38 & 52.33 & 49.45       \\
ACraft    & Gemini flash \cite{comanici2025gemini} & 57.28  & 50.83 & 49.17 & 52.65 & 52.48            \\
ACraft    & Llama 3 \cite{grattafiori2024llama3herdmodels}    & 49.04  & 48.70  & 45.32 & 43.95 & 46.75      \\
ACraft    & gpt-4o-mini \cite{openai2024gpt4technicalreport}      & 56.61  & 49.98  & 53.50 & 54.67 & 53.69   \\
ACraft    & Grok 3 \cite{grok3}    & 52.98  & 53.89  & 49.67 & 55.36 & 52.98      \\
\bottomrule
\end{tabular}%
}
\end{table}

\subsection{Results on FSCIL Benchmarks}
\noindent{\textbf{miniImageNet Dataset.}} As shown in Table \ref{table:comp_mini}, our proposed ACraft achieves the most destructive attack performance on the miniImageNet dataset, substantially outperforming all existing manually designed baselines. Specifically, compared with the original CLOSER model (62.78\% average accuracy), ACraft reduces the accuracy to only 23.10\%, resulting in an attack drop of 39.68 points, far exceeding traditional methods such as FGSM (4.77), PGD (4.76), and C\&W (1.68). Remarkably, ACraft not only induces the largest performance degradation across all incremental sessions but also maintains extremely consistent and low accuracies (approximately 12–23\%) throughout the entire incremental learning process. This demonstrates that ACraft can effectively destroy both base and novel class performance, overcoming the instability and inefficiency of expert-crafted attacks. These results clearly validate the superiority of our LLM-driven automatic attack generation framework, highlighting its capability to deliver both higher attack effectiveness and lower design cost in few-shot class-incremental learning scenarios. \textbf{In particular, ACraft is extremely efficient, only costing 47 minutes with gpt-4o-mini.}

\noindent{\textbf{CIFAR-100 and CUB200 Datasets.}} As shown in Fig. \ref{fig:cifar10} and \ref{fig:cub200}, ACraft exhibits outstanding attack performance on the CIFAR-100 and CUB200 datasets, further validating its robustness and generalization ability. Specifically, ACraft achieves the largest accuracy degradation across all incremental sessions, significantly outperforming classical handcrafted methods such as FGSM, PGD, and C\&W. Compared with the baseline CLOSER model, ACraft drastically reduces both the total and base accuracies while maintaining the lowest accuracy for new classes, indicating its ability to disrupt model learning consistently throughout the entire FSCIL process. Importantly, ACraft achieves these results without any manual design effort, highlighting the efficiency of our LLM-driven framework. These findings clearly demonstrate ACraft’s superior attack effectiveness and strong generalization efficiency across different datasets, establishing a new benchmark for automated adversarial attack discovery in FSCIL.

\subsection{Generalizability on more FSCIL Frameworks}
To further validate the generalizability of ACraft across different mainstream FSCIL frameworks, we conduct extensive experiments on multiple representative methods, including Limit~\cite{zhou2023few}, Approximation~\cite{wang2024approximation}, OrCo~\cite{ahmed2024orco}, and Tri-WE~\cite{zhou2023few}, on the miniImageNet dataset (see Table \ref{table:head_ablation}). From Table \ref{table:head_ablation}, we observe that ACraft consistently degrades the performance of all FSCIL methods by a large margin, confirming its strong generalization ability. For instance, when integrated with Limit, the average accuracy drops sharply from 67.96\% to 19.65\%. Similarly, Approximation, OrCo, and Tri-WE suffer significant performance declines from 65.88\% to 18.01\%, 67.38\% to 19.13\%, and 70.62\% to 19.56\%, respectively. These results demonstrate that ACraft retains strong generalizability and stability across different FSCIL methods, showing strong adaptability and transferability for diverse FSCIL frameworks.

\subsection{Ablation study}
\paragraph{The impact of various LLMs:} 
% To scrutinize the impact of various LLMs (i.e., GPT4o, Llama 4), we perform an ablation study of LLMs on the NAS-Bench-201 search space in the CIFAR-10 dataset (As shown Table \ref{tab:LLM} and the left panel of Fig. \ref{fig5}) and the TransNAS-Bench-101 Micro search space (as shown Fig. \ref{fig5} right). Specifically, we conduct experiments by averaging 4 independent runs to keep a fair comparison. From Table \ref{tab:LLM} and Fig. \ref{fig5}, we draw two conclusions: (1) Our ACraft demonstrates strong robustness across different LLMs. Specifically, ACraft achieves highly consistent results under different LLMs, showing minimal performance fluctuations across models. Moreover, ACraft is stable to noise caused by different training settings, further enhancing its reliability in real-world deployment.

To scrutinize the influence of different large language models (LLMs), we conduct an ablation study on six representative LLMs, i.e., GPT4o, Claude 3.7, DeepSeek V3, Gemini Flash, Llama 3, and Grok 3, using the CLOSER benchmark on miniImageNet (see Table \ref{tab:LLM}). Each result is averaged over four independent runs for fairness. As shown in Table \ref{tab:LLM}, ACraft demonstrates 
strong robustness across all LLMs, with “Attack Drop” values remaining consistent (46.75–53.69), indicating stable adaptation to diverse model behaviors. Notably, GPT4o achieves the best performance (53.69\%), reflecting its stronger reasoning and optimization capability in generating effective adversarial perturbations. Overall, ACraft maintains stable and reliable performance under different LLMs.

\subsection{Additional Evaluation}
Due to the page limit, this paper provides more experimental results in \textbf{App. \red{A}-\red{G}}.
 \ding{182} \noindent{\textbf{Detailed Prompt Engineering of ACraft}} are depicted in \textbf{App. \red{C}}. \ding{183} \noindent{\textbf{More ablation studies of ACraft}} are shown in \textbf{App. \red{D}}. \ding{184} \noindent{\textbf{Limitations and Discussion}} are presented in \textbf{App. \red{E}}.
 \ding{185} \noindent{\textbf{Visualizations of searched attack method}} are depicted in \textbf{App. \red{F}}.

 \noindent\textbf{Related Work.} The related work is provided in \textbf{App. \red{G}}.

\section{Conclusion and Future Works}\label{sec:conclusion}
In this work, we observe that existing attack methods in few-shot class-incremental learning (FSCIL) are manually designed, inefficient, and heavily dependent on expert knowledge. To address this limitation, we propose ACraft, an LLM-driven framework that automatically discovers effective adversarial attack algorithms without human intervention. By incorporating a Proximal Policy Optimization (PPO)-based reinforcement learning strategy, ACraft iteratively refines prompts to generate stronger and more adaptive attacks against FSCIL models. The proposed method significantly outperforms expert-designed attacks while maintaining the lowest computational cost on multiple FSCIL benchmarks. We believe ACraft introduces a novel paradigm for automated adversarial attack generation and will inspire future research on secure and robust continual learning systems. In future work, we plan to further enhance ACraft’s reasoning and adaptation capabilities for broader applications.
% \input{sec/X_suppl} % 补充材料通常不放在正文里，或者需要另起一页

% --- 参考文献 ---
{
    \small
    % [注意] 必须确保目录下有 ieeenat_fullname.bst 文件
    % 如果没有，请改用 \bibliographystyle{plainnat} 或 \bibliographystyle{unsrt}
    \bibliographystyle{ieeenat_fullname}
    \bibliography{main}

@String(ECCV= {Eur. Conf. Comput. Vis.})

@String(ECCV  = {ECCV})

@article{r6,
  title={Continual learning: Applications and the road forward},
  author={Verwimp, Eli and Aljundi, Rahaf and Ben-David, Shai and Bethge, Matthias and Cossu, Andrea and Gepperth, Alexander and Hayes, Tyler L and H{\"u}llermeier, Eyke and Kanan, Christopher and Kudithipudi, Dhireesha and others},
  journal={arXiv preprint arXiv:2311.11908},
  year={2023}
}

@inproceedings{r7,
  title={Attraction Diminishing and Distributing for Few-Shot Class-Incremental Learning},
  author={Zhao, Li-Jun and Chen, Zhen-Duo and Wang, Yongxin and Luo, Xin and Xu, Xin-Shun},
  booktitle={Proceedings of the Computer Vision and Pattern Recognition Conference},
  pages={25657--25666},
  year={2025}
}

@inproceedings{r8,
  title={SEC-Prompt: SEmantic Complementary Prompting for Few-Shot Class-Incremental Learning},
  author={Liu, Ye and Yang, Meng},
  booktitle={Proceedings of the Computer Vision and Pattern Recognition Conference},
  pages={25643--25656},
  year={2025}
}

@inproceedings{r9,
  title={Enhancing Few-Shot Class-Incremental Learning via Training-Free Bi-Level Modality Calibration},
  author={Chen, Yiyang and Ding, Tianyu and Wang, Lei and Huo, Jing and Gao, Yang and Li, Wenbin},
  booktitle={Proceedings of the Computer Vision and Pattern Recognition Conference},
  pages={9881--9890},
  year={2025}
}

@article{r10,
  title={An efficient memory module for graph few-shot class-incremental learning},
  author={Li, Dong and Zhang, Aijia and Gao, Junqi and Qi, Biqing},
  journal={Advances in Neural Information Processing Systems},
  volume={37},
  pages={130084--130108},
  year={2024}
}

@inproceedings{r11,
  title={Closer: Towards better representation learning for few-shot class-incremental learning},
  author={Oh, Junghun and Baik, Sungyong and Lee, Kyoung Mu},
  booktitle={European Conference on Computer Vision},
  pages={18--35},
  year={2024},
  organization={Springer}
}

@article{CoT,
  title={Chain-of-thought prompting elicits reasoning in large language models},
  author={Wei, Jason and Wang, Xuezhi and Schuurmans, Dale and Bosma, Maarten and Xia, Fei and Chi, Ed and Le, Quoc V and Zhou, Denny and others},
  journal={Advances in neural information processing systems},
  volume={35},
  pages={24824--24837},
  year={2022}
}

@article{r17,
  title={A continual learning survey: Defying forgetting in classification tasks},
  author={De Lange, Matthias and Aljundi, Rahaf and Masana, Marc and Parisot, Sarah and Jia, Xu and Leonardis, Ale{\v{s}} and Slabaugh, Gregory and Tuytelaars, Tinne},
  journal={IEEE transactions on pattern analysis and machine intelligence},
  volume={44},
  number={7},
  pages={3366--3385},
  year={2021},
  publisher={IEEE}
}

@article{r18,
  title={Few-shot adversarial prompt learning on vision-language models},
  author={Zhou, Yiwei and Xia, Xiaobo and Lin, Zhiwei and Han, Bo and Liu, Tongliang},
  journal={Advances in Neural Information Processing Systems},
  volume={37},
  pages={3122--3156},
  year={2024}
}

@article{r19,
  title={Adversarial robustness of prompt-based few-shot learning for natural language understanding},
  author={Nookala, Venkata Prabhakara Sarath and Verma, Gaurav and Mukherjee, Subhabrata and Kumar, Srijan},
  journal={arXiv preprint arXiv:2306.11066},
  year={2023}
}

@article{r20,
  title={Learning domain invariant prompt for vision-language models},
  author={Zhao, Cairong and Wang, Yubin and Jiang, Xinyang and Shen, Yifei and Song, Kaitao and Li, Dongsheng and Miao, Duoqian},
  journal={IEEE Transactions on Image Processing},
  volume={33},
  pages={1348--1360},
  year={2024},
  publisher={IEEE}
}

@inproceedings{oh2024closer,
  title={CLOSER: Towards Better Representation Learning for Few-Shot Class-Incremental Learning},
  author={Oh, Junghun and Baik, Sungyong and Lee, Kyoung Mu},
  booktitle=ECCV,
  year={2024}
}

@inproceedings{matchingnet,
  author    = {Oriol Vinyals and
               Charles Blundell and
               Tim Lillicrap and
               Koray Kavukcuoglu and
               Daan Wierstra},
  title     = {Matching Networks for One Shot Learning},
  booktitle = NeurIPS,
  year      = {2016}
}

@article{krizhevsky2009learning,
  title={Learning multiple layers of features from tiny images},
  author={Krizhevsky, Alex and Hinton, Geoffrey and others},
  year={2009},
  publisher={Toronto, ON, Canada}
}

@article{wah2011caltech,
  title={The caltech-ucsd birds-200-2011 dataset},
  author={Wah, Catherine and Branson, Steve and Welinder, Peter and Perona, Pietro and Belongie, Serge},
  year={2011},
  publisher={California Institute of Technology}
}

@inproceedings{wang2024approximation,
  title={On the approximation risk of few-shot class-incremental learning},
  author={Wang, Xuan and Ji, Zhong and Liu, Xiyao and Pang, Yanwei and Han, Jungong},
  booktitle={European Conference on Computer Vision},
  pages={162--178},
  year={2024},
  organization={Springer}
}

@inproceedings{ahmed2024orco,
    title={OrCo: Towards Better Generalization via Orthogonality and Contrast for Few-Shot Class-Incremental Learning},
    author={Ahmed, Noor and Kukleva, Anna and Schiele, Bernt},
    booktitle={41st IEEE/CVF Conference on Computer Vision and Pattern Recognition},
    year={2024},
    organization={IEEE}
}

@ARTICLE{zhou2023few,
    author={Zhou, Da-Wei and Ye, Han-Jia and Ma, Liang and Xie, Di and Pu, Shiliang and Zhan, De-Chuan},
    journal={IEEE Transactions on Pattern Analysis and Machine Intelligence}, 
    title={Few-Shot Class-Incremental Learning by Sampling Multi-Phase Tasks}, 
    year={2023},
    volume={45},
    number={11},
    pages={12816-12831},
    doi={10.1109/TPAMI.2022.3200865}
}

@article{madry2017towards,
  title={Towards deep learning models resistant to adversarial attacks},
  author={Madry, Aleksander and Makelov, Aleksandar and Schmidt, Ludwig and Tsipras, Dimitris and Vladu, Adrian},
  journal={arXiv preprint arXiv:1706.06083},
  year={2017}
}

@inproceedings{carlini2017towards,
  title={Towards evaluating the robustness of neural networks},
  author={Carlini, Nicholas and Wagner, David},
  booktitle={2017 ieee symposium on security and privacy (sp)},
  pages={39--57},
  year={2017},
  organization={Ieee}
}

@article{goodfellow2014explaining,
  title={Explaining and harnessing adversarial examples},
  author={Goodfellow, Ian J and Shlens, Jonathon and Szegedy, Christian},
  journal={arXiv preprint arXiv:1412.6572},
  year={2014}
}

@inproceedings{moosavi2016deepfool,
  title={Deepfool: a simple and accurate method to fool deep neural networks},
  author={Moosavi-Dezfooli, Seyed-Mohsen and Fawzi, Alhussein and Frossard, Pascal},
  booktitle={Proceedings of the IEEE conference on computer vision and pattern recognition},
  pages={2574--2582},
  year={2016}
}

@article{achiam2023gpt,
  title={Gpt-4 technical report},
  author={Achiam, Josh and Adler, Steven and Agarwal, Sandhini and Ahmad, Lama and Akkaya, Ilge and Aleman, Florencia Leoni and Almeida, Diogo and Altenschmidt, Janko and Altman, Sam and Anadkat, Shyamal and others},
  journal={arXiv preprint arXiv:2303.08774},
  year={2023}
}

@inproceedings{TheC3,
  title={The Claude 3 Model Family: Opus, Sonnet, Haiku},
  author={},
  url={https://api.semanticscholar.org/CorpusID:268232499}
}

@article{liu2024deepseek,
  title={Deepseek-v3 technical report},
  author={Liu, Aixin and Feng, Bei and Xue, Bing and Wang, Bingxuan and Wu, Bochao and Lu, Chengda and Zhao, Chenggang and Deng, Chengqi and Zhang, Chenyu and Ruan, Chong and others},
  journal={arXiv preprint arXiv:2412.19437},
  year={2024}
}

@article{comanici2025gemini,
  title={Gemini 2.5: Pushing the frontier with advanced reasoning, multimodality, long context, and next generation agentic capabilities},
  author={Comanici, Gheorghe and Bieber, Eric and Schaekermann, Mike and Pasupat, Ice and Sachdeva, Noveen and Dhillon, Inderjit and Blistein, Marcel and Ram, Ori and Zhang, Dan and Rosen, Evan and others},
  journal={arXiv preprint arXiv:2507.06261},
  year={2025}
}

@misc{grattafiori2024llama3herdmodels,
      title={The Llama 3 Herd of Models}, 
      author={Aaron Grattafiori and Abhimanyu Dubey and Abhinav Jauhri and Abhinav Pandey and Abhishek Kadian and Ahmad Al-Dahle and Aiesha Letman and Akhil Mathur and Alan Schelten and Alex Vaughan and Amy Yang and Angela Fan and Anirudh Goyal and Anthony Hartshorn and Aobo Yang and Archi Mitra and Archie Sravankumar and Artem Korenev and Arthur Hinsvark and Arun Rao and Aston Zhang and Aurelien Rodriguez and Austen Gregerson and Ava Spataru and Baptiste Roziere and Bethany Biron and Binh Tang and Bobbie Chern and Charlotte Caucheteux and Chaya Nayak and Chloe Bi and Chris Marra and Chris McConnell and Christian Keller and Christophe Touret and Chunyang Wu and Corinne Wong and Cristian Canton Ferrer and Cyrus Nikolaidis and Damien Allonsius and Daniel Song and Danielle Pintz and Danny Livshits and Danny Wyatt and David Esiobu and Dhruv Choudhary and Dhruv Mahajan and Diego Garcia-Olano and Diego Perino and Dieuwke Hupkes and Egor Lakomkin and Ehab AlBadawy and Elina Lobanova and Emily Dinan and Eric Michael Smith and Filip Radenovic and Francisco Guzmán and Frank Zhang and Gabriel Synnaeve and Gabrielle Lee and Georgia Lewis Anderson and Govind Thattai and Graeme Nail and Gregoire Mialon and Guan Pang and Guillem Cucurell and Hailey Nguyen and Hannah Korevaar and Hu Xu and Hugo Touvron and Iliyan Zarov and Imanol Arrieta Ibarra and Isabel Kloumann and Ishan Misra and Ivan Evtimov and Jack Zhang and Jade Copet and Jaewon Lee and Jan Geffert and Jana Vranes and Jason Park and Jay Mahadeokar and Jeet Shah and Jelmer van der Linde and Jennifer Billock and Jenny Hong and Jenya Lee and Jeremy Fu and Jianfeng Chi and Jianyu Huang and Jiawen Liu and Jie Wang and Jiecao Yu and Joanna Bitton and Joe Spisak and Jongsoo Park and Joseph Rocca and Joshua Johnstun and Joshua Saxe and Junteng Jia and Kalyan Vasuden Alwala and Karthik Prasad and Kartikeya Upasani and Kate Plawiak and Ke Li and Kenneth Heafield and Kevin Stone and Khalid El-Arini and Krithika Iyer and Kshitiz Malik and Kuenley Chiu and Kunal Bhalla and Kushal Lakhotia and Lauren Rantala-Yeary and Laurens van der Maaten and Lawrence Chen and Liang Tan and Liz Jenkins and Louis Martin and Lovish Madaan and Lubo Malo and Lukas Blecher and Lukas Landzaat and Luke de Oliveira and Madeline Muzzi and Mahesh Pasupuleti and Mannat Singh and Manohar Paluri and Marcin Kardas and Maria Tsimpoukelli and Mathew Oldham and Mathieu Rita and Maya Pavlova and Melanie Kambadur and Mike Lewis and Min Si and Mitesh Kumar Singh and Mona Hassan and Naman Goyal and Narjes Torabi and Nikolay Bashlykov and Nikolay Bogoychev and Niladri Chatterji and Ning Zhang and Olivier Duchenne and Onur Çelebi and Patrick Alrassy and Pengchuan Zhang and Pengwei Li and Petar Vasic and Peter Weng and Prajjwal Bhargava and Pratik Dubal and Praveen Krishnan and Punit Singh Koura and Puxin Xu and Qing He and Qingxiao Dong and Ragavan Srinivasan and Raj Ganapathy and Ramon Calderer and Ricardo Silveira Cabral and Robert Stojnic and Roberta Raileanu and Rohan Maheswari and Rohit Girdhar and Rohit Patel and Romain Sauvestre and Ronnie Polidoro and Roshan Sumbaly and Ross Taylor and Ruan Silva and Rui Hou and Rui Wang and Saghar Hosseini and Sahana Chennabasappa and Sanjay Singh and Sean Bell and Seohyun Sonia Kim and Sergey Edunov and Shaoliang Nie and Sharan Narang and Sharath Raparthy and Sheng Shen and Shengye Wan and Shruti Bhosale and Shun Zhang and Simon Vandenhende and Soumya Batra and Spencer Whitman and Sten Sootla and Stephane Collot and Suchin Gururangan and Sydney Borodinsky and Tamar Herman and Tara Fowler and Tarek Sheasha and Thomas Georgiou and Thomas Scialom and Tobias Speckbacher and Todor Mihaylov and Tong Xiao and Ujjwal Karn and Vedanuj Goswami and Vibhor Gupta and Vignesh Ramanathan and Viktor Kerkez and Vincent Gonguet and Virginie Do and Vish Vogeti and Vítor Albiero and Vladan Petrovic and Weiwei Chu and Wenhan Xiong and Wenyin Fu and Whitney Meers and Xavier Martinet and Xiaodong Wang and Xiaofang Wang and Xiaoqing Ellen Tan and Xide Xia and Xinfeng Xie and Xuchao Jia and Xuewei Wang and Yaelle Goldschlag and Yashesh Gaur and Yasmine Babaei and Yi Wen and Yiwen Song and Yuchen Zhang and Yue Li and Yuning Mao and Zacharie Delpierre Coudert and Zheng Yan and Zhengxing Chen and Zoe Papakipos and Aaditya Singh and Aayushi Srivastava and Abha Jain and Adam Kelsey and Adam Shajnfeld and Adithya Gangidi and Adolfo Victoria and Ahuva Goldstand and Ajay Menon and Ajay Sharma and Alex Boesenberg and Alexei Baevski and Allie Feinstein and Amanda Kallet and Amit Sangani and Amos Teo and Anam Yunus and Andrei Lupu and Andres Alvarado and Andrew Caples and Andrew Gu and Andrew Ho and Andrew Poulton and Andrew Ryan and Ankit Ramchandani and Annie Dong and Annie Franco and Anuj Goyal and Aparajita Saraf and Arkabandhu Chowdhury and Ashley Gabriel and Ashwin Bharambe and Assaf Eisenman and Azadeh Yazdan and Beau James and Ben Maurer and Benjamin Leonhardi and Bernie Huang and Beth Loyd and Beto De Paola and Bhargavi Paranjape and Bing Liu and Bo Wu and Boyu Ni and Braden Hancock and Bram Wasti and Brandon Spence and Brani Stojkovic and Brian Gamido and Britt Montalvo and Carl Parker and Carly Burton and Catalina Mejia and Ce Liu and Changhan Wang and Changkyu Kim and Chao Zhou and Chester Hu and Ching-Hsiang Chu and Chris Cai and Chris Tindal and Christoph Feichtenhofer and Cynthia Gao and Damon Civin and Dana Beaty and Daniel Kreymer and Daniel Li and David Adkins and David Xu and Davide Testuggine and Delia David and Devi Parikh and Diana Liskovich and Didem Foss and Dingkang Wang and Duc Le and Dustin Holland and Edward Dowling and Eissa Jamil and Elaine Montgomery and Eleonora Presani and Emily Hahn and Emily Wood and Eric-Tuan Le and Erik Brinkman and Esteban Arcaute and Evan Dunbar and Evan Smothers and Fei Sun and Felix Kreuk and Feng Tian and Filippos Kokkinos and Firat Ozgenel and Francesco Caggioni and Frank Kanayet and Frank Seide and Gabriela Medina Florez and Gabriella Schwarz and Gada Badeer and Georgia Swee and Gil Halpern and Grant Herman and Grigory Sizov and Guangyi and Zhang and Guna Lakshminarayanan and Hakan Inan and Hamid Shojanazeri and Han Zou and Hannah Wang and Hanwen Zha and Haroun Habeeb and Harrison Rudolph and Helen Suk and Henry Aspegren and Hunter Goldman and Hongyuan Zhan and Ibrahim Damlaj and Igor Molybog and Igor Tufanov and Ilias Leontiadis and Irina-Elena Veliche and Itai Gat and Jake Weissman and James Geboski and James Kohli and Janice Lam and Japhet Asher and Jean-Baptiste Gaya and Jeff Marcus and Jeff Tang and Jennifer Chan and Jenny Zhen and Jeremy Reizenstein and Jeremy Teboul and Jessica Zhong and Jian Jin and Jingyi Yang and Joe Cummings and Jon Carvill and Jon Shepard and Jonathan McPhie and Jonathan Torres and Josh Ginsburg and Junjie Wang and Kai Wu and Kam Hou U and Karan Saxena and Kartikay Khandelwal and Katayoun Zand and Kathy Matosich and Kaushik Veeraraghavan and Kelly Michelena and Keqian Li and Kiran Jagadeesh and Kun Huang and Kunal Chawla and Kyle Huang and Lailin Chen and Lakshya Garg and Lavender A and Leandro Silva and Lee Bell and Lei Zhang and Liangpeng Guo and Licheng Yu and Liron Moshkovich and Luca Wehrstedt and Madian Khabsa and Manav Avalani and Manish Bhatt and Martynas Mankus and Matan Hasson and Matthew Lennie and Matthias Reso and Maxim Groshev and Maxim Naumov and Maya Lathi and Meghan Keneally and Miao Liu and Michael L. Seltzer and Michal Valko and Michelle Restrepo and Mihir Patel and Mik Vyatskov and Mikayel Samvelyan and Mike Clark and Mike Macey and Mike Wang and Miquel Jubert Hermoso and Mo Metanat and Mohammad Rastegari and Munish Bansal and Nandhini Santhanam and Natascha Parks and Natasha White and Navyata Bawa and Nayan Singhal and Nick Egebo and Nicolas Usunier and Nikhil Mehta and Nikolay Pavlovich Laptev and Ning Dong and Norman Cheng and Oleg Chernoguz and Olivia Hart and Omkar Salpekar and Ozlem Kalinli and Parkin Kent and Parth Parekh and Paul Saab and Pavan Balaji and Pedro Rittner and Philip Bontrager and Pierre Roux and Piotr Dollar and Polina Zvyagina and Prashant Ratanchandani and Pritish Yuvraj and Qian Liang and Rachad Alao and Rachel Rodriguez and Rafi Ayub and Raghotham Murthy and Raghu Nayani and Rahul Mitra and Rangaprabhu Parthasarathy and Raymond Li and Rebekkah Hogan and Robin Battey and Rocky Wang and Russ Howes and Ruty Rinott and Sachin Mehta and Sachin Siby and Sai Jayesh Bondu and Samyak Datta and Sara Chugh and Sara Hunt and Sargun Dhillon and Sasha Sidorov and Satadru Pan and Saurabh Mahajan and Saurabh Verma and Seiji Yamamoto and Sharadh Ramaswamy and Shaun Lindsay and Shaun Lindsay and Sheng Feng and Shenghao Lin and Shengxin Cindy Zha and Shishir Patil and Shiva Shankar and Shuqiang Zhang and Shuqiang Zhang and Sinong Wang and Sneha Agarwal and Soji Sajuyigbe and Soumith Chintala and Stephanie Max and Stephen Chen and Steve Kehoe and Steve Satterfield and Sudarshan Govindaprasad and Sumit Gupta and Summer Deng and Sungmin Cho and Sunny Virk and Suraj Subramanian and Sy Choudhury and Sydney Goldman and Tal Remez and Tamar Glaser and Tamara Best and Thilo Koehler and Thomas Robinson and Tianhe Li and Tianjun Zhang and Tim Matthews and Timothy Chou and Tzook Shaked and Varun Vontimitta and Victoria Ajayi and Victoria Montanez and Vijai Mohan and Vinay Satish Kumar and Vishal Mangla and Vlad Ionescu and Vlad Poenaru and Vlad Tiberiu Mihailescu and Vladimir Ivanov and Wei Li and Wenchen Wang and Wenwen Jiang and Wes Bouaziz and Will Constable and Xiaocheng Tang and Xiaojian Wu and Xiaolan Wang and Xilun Wu and Xinbo Gao and Yaniv Kleinman and Yanjun Chen and Ye Hu and Ye Jia and Ye Qi and Yenda Li and Yilin Zhang and Ying Zhang and Yossi Adi and Youngjin Nam and Yu and Wang and Yu Zhao and Yuchen Hao and Yundi Qian and Yunlu Li and Yuzi He and Zach Rait and Zachary DeVito and Zef Rosnbrick and Zhaoduo Wen and Zhenyu Yang and Zhiwei Zhao and Zhiyu Ma},
      year={2024},
      eprint={2407.21783},
      archivePrefix={arXiv},
      primaryClass={cs.AI},
      url={https://arxiv.org/abs/2407.21783}, 
}

@misc{openai2024gpt4technicalreport,
      title={GPT-4 Technical Report}, 
      author={OpenAI and Josh Achiam and Steven Adler and Sandhini Agarwal and Lama Ahmad and Ilge Akkaya and Florencia Leoni Aleman and Diogo Almeida and Janko Altenschmidt and Sam Altman and Shyamal Anadkat and Red Avila and Igor Babuschkin and Suchir Balaji and Valerie Balcom and Paul Baltescu and Haiming Bao and Mohammad Bavarian and Jeff Belgum and Irwan Bello and Jake Berdine and Gabriel Bernadett-Shapiro and Christopher Berner and Lenny Bogdonoff and Oleg Boiko and Madelaine Boyd and Anna-Luisa Brakman and Greg Brockman and Tim Brooks and Miles Brundage and Kevin Button and Trevor Cai and Rosie Campbell and Andrew Cann and Brittany Carey and Chelsea Carlson and Rory Carmichael and Brooke Chan and Che Chang and Fotis Chantzis and Derek Chen and Sully Chen and Ruby Chen and Jason Chen and Mark Chen and Ben Chess and Chester Cho and Casey Chu and Hyung Won Chung and Dave Cummings and Jeremiah Currier and Yunxing Dai and Cory Decareaux and Thomas Degry and Noah Deutsch and Damien Deville and Arka Dhar and David Dohan and Steve Dowling and Sheila Dunning and Adrien Ecoffet and Atty Eleti and Tyna Eloundou and David Farhi and Liam Fedus and Niko Felix and Simón Posada Fishman and Juston Forte and Isabella Fulford and Leo Gao and Elie Georges and Christian Gibson and Vik Goel and Tarun Gogineni and Gabriel Goh and Rapha Gontijo-Lopes and Jonathan Gordon and Morgan Grafstein and Scott Gray and Ryan Greene and Joshua Gross and Shixiang Shane Gu and Yufei Guo and Chris Hallacy and Jesse Han and Jeff Harris and Yuchen He and Mike Heaton and Johannes Heidecke and Chris Hesse and Alan Hickey and Wade Hickey and Peter Hoeschele and Brandon Houghton and Kenny Hsu and Shengli Hu and Xin Hu and Joost Huizinga and Shantanu Jain and Shawn Jain and Joanne Jang and Angela Jiang and Roger Jiang and Haozhun Jin and Denny Jin and Shino Jomoto and Billie Jonn and Heewoo Jun and Tomer Kaftan and Łukasz Kaiser and Ali Kamali and Ingmar Kanitscheider and Nitish Shirish Keskar and Tabarak Khan and Logan Kilpatrick and Jong Wook Kim and Christina Kim and Yongjik Kim and Jan Hendrik Kirchner and Jamie Kiros and Matt Knight and Daniel Kokotajlo and Łukasz Kondraciuk and Andrew Kondrich and Aris Konstantinidis and Kyle Kosic and Gretchen Krueger and Vishal Kuo and Michael Lampe and Ikai Lan and Teddy Lee and Jan Leike and Jade Leung and Daniel Levy and Chak Ming Li and Rachel Lim and Molly Lin and Stephanie Lin and Mateusz Litwin and Theresa Lopez and Ryan Lowe and Patricia Lue and Anna Makanju and Kim Malfacini and Sam Manning and Todor Markov and Yaniv Markovski and Bianca Martin and Katie Mayer and Andrew Mayne and Bob McGrew and Scott Mayer McKinney and Christine McLeavey and Paul McMillan and Jake McNeil and David Medina and Aalok Mehta and Jacob Menick and Luke Metz and Andrey Mishchenko and Pamela Mishkin and Vinnie Monaco and Evan Morikawa and Daniel Mossing and Tong Mu and Mira Murati and Oleg Murk and David Mély and Ashvin Nair and Reiichiro Nakano and Rajeev Nayak and Arvind Neelakantan and Richard Ngo and Hyeonwoo Noh and Long Ouyang and Cullen O'Keefe and Jakub Pachocki and Alex Paino and Joe Palermo and Ashley Pantuliano and Giambattista Parascandolo and Joel Parish and Emy Parparita and Alex Passos and Mikhail Pavlov and Andrew Peng and Adam Perelman and Filipe de Avila Belbute Peres and Michael Petrov and Henrique Ponde de Oliveira Pinto and Michael and Pokorny and Michelle Pokrass and Vitchyr H. Pong and Tolly Powell and Alethea Power and Boris Power and Elizabeth Proehl and Raul Puri and Alec Radford and Jack Rae and Aditya Ramesh and Cameron Raymond and Francis Real and Kendra Rimbach and Carl Ross and Bob Rotsted and Henri Roussez and Nick Ryder and Mario Saltarelli and Ted Sanders and Shibani Santurkar and Girish Sastry and Heather Schmidt and David Schnurr and John Schulman and Daniel Selsam and Kyla Sheppard and Toki Sherbakov and Jessica Shieh and Sarah Shoker and Pranav Shyam and Szymon Sidor and Eric Sigler and Maddie Simens and Jordan Sitkin and Katarina Slama and Ian Sohl and Benjamin Sokolowsky and Yang Song and Natalie Staudacher and Felipe Petroski Such and Natalie Summers and Ilya Sutskever and Jie Tang and Nikolas Tezak and Madeleine B. Thompson and Phil Tillet and Amin Tootoonchian and Elizabeth Tseng and Preston Tuggle and Nick Turley and Jerry Tworek and Juan Felipe Cerón Uribe and Andrea Vallone and Arun Vijayvergiya and Chelsea Voss and Carroll Wainwright and Justin Jay Wang and Alvin Wang and Ben Wang and Jonathan Ward and Jason Wei and CJ Weinmann and Akila Welihinda and Peter Welinder and Jiayi Weng and Lilian Weng and Matt Wiethoff and Dave Willner and Clemens Winter and Samuel Wolrich and Hannah Wong and Lauren Workman and Sherwin Wu and Jeff Wu and Michael Wu and Kai Xiao and Tao Xu and Sarah Yoo and Kevin Yu and Qiming Yuan and Wojciech Zaremba and Rowan Zellers and Chong Zhang and Marvin Zhang and Shengjia Zhao and Tianhao Zheng and Juntang Zhuang and William Zhuk and Barret Zoph},
      year={2024},
      eprint={2303.08774},
      archivePrefix={arXiv},
      primaryClass={cs.CL},
      url={https://arxiv.org/abs/2303.08774}, 
}

@misc{grok3,
    title={grok-3},
    url={https://grok.com/}
}
}

\end{document}